\PassOptionsToPackage{unicode}{hyperref}
\PassOptionsToPackage{hyphens}{url}
\PassOptionsToPackage{dvipsnames,svgnames,x11names}{xcolor}
\documentclass[
]{article}
\usepackage{xcolor}
\usepackage[margin=1in]{geometry}
\usepackage{amsmath,amssymb}
\usepackage{iftex}
\ifPDFTeX
  \usepackage[T1]{fontenc}
  \usepackage[utf8]{inputenc}
  \usepackage{textcomp} 
\else 
  \usepackage{unicode-math} 
  \defaultfontfeatures{Scale=MatchLowercase}
  \defaultfontfeatures[\rmfamily]{Ligatures=TeX,Scale=1}
\fi
\usepackage{lmodern}
\ifPDFTeX\else
\fi
\IfFileExists{upquote.sty}{\usepackage{upquote}}{}
\IfFileExists{microtype.sty}{
  \usepackage[]{microtype}
  \UseMicrotypeSet[protrusion]{basicmath} 
}{}
\makeatletter
\@ifundefined{KOMAClassName}{
  \IfFileExists{parskip.sty}{%
    \usepackage{parskip}
  }{
    \setlength{\parindent}{0pt}
    \setlength{\parskip}{6pt plus 2pt minus 1pt}}
}{
  \KOMAoptions{parskip=half}}
\makeatother
\usepackage{longtable,booktabs,array}
\usepackage{calc} 
\usepackage{etoolbox}
\makeatletter
\patchcmd\longtable{\par}{\if@noskipsec\mbox{}\fi\par}{}{}
\makeatother
\IfFileExists{footnotehyper.sty}{\usepackage{footnotehyper}}{\usepackage{footnote}}
\makesavenoteenv{longtable}
\usepackage{graphicx}
\makeatletter
\newsavebox\pandoc@box
\newcommand*\pandocbounded[1]{
  \sbox\pandoc@box{#1}%
  \Gscale@div\@tempa{\textheight}{\dimexpr\ht\pandoc@box+\dp\pandoc@box\relax}%
  \Gscale@div\@tempb{\linewidth}{\wd\pandoc@box}%
  \ifdim\@tempb\p@<\@tempa\p@\let\@tempa\@tempb\fi
  \ifdim\@tempa\p@<\p@\scalebox{\@tempa}{\usebox\pandoc@box}%
  \else\usebox{\pandoc@box}%
  \fi%
}
\def\fps@figure{htbp}
\makeatother
\providecommand{\tightlist}{%
  \setlength{\itemsep}{0pt}\setlength{\parskip}{0pt}}
\usepackage{bookmark}
\IfFileExists{xurl.sty}{\usepackage{xurl}}{} 
\hypersetup{
  colorlinks=true,
  linkcolor={NavyBlue},
  filecolor={Maroon},
  citecolor={Blue},
  urlcolor={NavyBlue},
  pdfcreator={LaTeX via pandoc}}

\author{}
\date{}

\begin{document}

{
\setcounter{tocdepth}{2}
\tableofcontents
}
\section{OCRR: A Benchmark for Online Correction Recovery under
Distribution
Shift}\label{ocrr-a-benchmark-for-online-correction-recovery-under-distribution-shift}

\textbf{Adrian Grassi}\\
Independent Researcher\\
\href{mailto:adriangrassi@gmail.com}{\nolinkurl{adriangrassi@gmail.com}}\\
ORCID:
\href{https://orcid.org/0009-0007-4890-5393}{0009-0007-4890-5393}\\
Code:
\href{https://github.com/adriangrassi/ocrr-benchmark}{github.com/adriangrassi/ocrr-benchmark}

\begin{center}\rule{0.5\linewidth}{0.5pt}\end{center}

\subsection{Abstract}\label{abstract}

Static benchmarks measure a model frozen at training time. Real systems
face distribution shift --- new categories, paraphrased queries, drift
--- and must recover \emph{online} via user corrections. No existing
benchmark measures recovery speed under correction streams. We introduce
\textbf{OCRR (Online Correction Recovery Rate)}: a benchmark that
streams a corpus through a classification system, applies oracle or
stochastic corrections to wrong predictions, and reports two curves:
novel-class accuracy and original-distribution accuracy versus
correction count. We evaluate the substrate alongside nine baseline
algorithms from five families plus seven bounded-storage variants of the
substrate for the Pareto sweep, including standard online-learning
baselines (\texttt{river}), continual-learning methods (EWC, A-GEM,
LwF), retrieval/parametric hybrids (kNN-LM), parameter-efficient
fine-tuning of a 1.5 B-parameter encoder (LoRA on DeBERTa-v3-large), and
a hash-chained append-only substrate (\emph{Substrate}). On Banking77
and CLINC150, under oracle and sparse correction policies, the substrate
is the only system that simultaneously recovers novel-class accuracy
(88.7 ± 2.9 \%) and retains original-distribution accuracy (95.4 ± 0.8
\%) --- beating the next-best published continual-learning baseline by
32.6 percentage points at equal memory budget, and beating
LoRA-on-DeBERTa-v3-large by 84.6 percentage points on retention. We
further find that classification accuracy remains stable at 99 \% even
as approximate-nearest-neighbour recall@5 degrades from 0.69 to 0.23
across 10 k → 10 M corpus scales, suggesting the substrate's margin-band
majority vote is robust to retrieval imperfection in a way that pure
top-k recall metrics do not predict. Code and data are available at
\href{https://github.com/adriangrassi/ocrr-benchmark}{github.com/adriangrassi/ocrr-benchmark}.

\begin{center}\rule{0.5\linewidth}{0.5pt}\end{center}

\subsection{1. Introduction}\label{introduction}

When a deployed classifier makes a mistake, the practitioner has three
choices: ignore it, queue corrections for the next retrain cycle (hours
to days), or update the model online. The third option is what
production systems actually need --- the data has shifted, the user
already knows the right answer, and waiting until Tuesday's retrain is
unacceptable.

Yet there is no benchmark for \emph{how well} a classifier recovers when
corrections arrive online. Banking77 (Casanueva et al.~2020), GLUE (Wang
et al.~2019), and MMLU (Hendrycks et al.~2021) all measure a model
frozen at training time. The continual-learning community has
Permuted-MNIST, Split-CIFAR-100, and incremental-class CIFAR (Lopez-Paz
\& Ranzato 2017; van de Ven \& Tolias 2019), but those benchmarks
evaluate models \emph{after} a full task has been seen --- not during a
stream of user corrections.

We argue that \emph{correction recovery rate} is a property that is
critical for deployed systems but inadequately captured by current
benchmarks. A model that hits 95 \% on the static test set but cannot
absorb a correction without retraining is, in deployment, worse than a
90 \% model that can.

This paper contributes:

\begin{enumerate}
\def\labelenumi{\arabic{enumi}.}
\tightlist
\item
  \textbf{OCRR}, a benchmark for online correction recovery. We define a
  stream- based protocol, two evaluation axes (novel and original),
  three correction policies (oracle, random-50 \%, random-10 \%), and
  three storage budgets (unbounded, bounded reservoir, bounded FIFO).
\item
  \textbf{Nine baseline algorithms} spanning five families: three
  strawmen (static-kNN, static-linear, online-linear), three
  continual-learning methods (EWC, A-GEM, LwF), one online-ML library
  (\texttt{river}), one retrieval/parametric hybrid (kNN-LM), and one
  parameter-efficient fine-tune (LoRA on DeBERTa-v3-large). We benchmark
  all nine against our substrate, a hash-chained append-only ledger with
  margin-band majority voting.
\item
  \textbf{162 system runs} in the main full sweep across 2 datasets × 3
  policies × 3 seeds × 9 systems, plus a separate 36-run
  storage-vs-recovery Pareto sweep over seven bounded-substrate variants
  (4 reservoir + 3 FIFO budget points), plus a 3-seed LoRA-DeBERTa cell.
\item
  \textbf{A clean characterisation} of where each method sits on the
  storage-vs-forgetting Pareto, exposing structural trade-offs that
  static benchmarks hide.
\end{enumerate}

\begin{center}\rule{0.5\linewidth}{0.5pt}\end{center}

\subsection{2. Background and Related
Work}\label{background-and-related-work}

\subsubsection{2.1 Static benchmarks}\label{static-benchmarks}

Banking77 (PolyAI, Casanueva et al.~2020) is a 77-intent classification
benchmark widely used in dialogue systems. CLINC150 (Larson et al.~2019)
extends to 151 intents across 10 domains. Both report a single accuracy
on a frozen test split. Recent leaderboards (Cohere, OpenAI) report
cascade ensembles reaching 95--96 \% on Banking77, but the operating
point itself is static.

\subsubsection{2.2 Online learning}\label{online-learning}

Classical online-ML libraries (\texttt{river}, the active fork of
\texttt{scikit-multiflow}, Montiel et al.~2018) assume each example
arrives as a stream and the model is updated incrementally. The
constraint is ``no historical-data storage.'' Algorithms include online
logistic regression, Hoeffding trees, and adaptive random forests.

\subsubsection{2.3 Continual learning}\label{continual-learning}

The continual-learning literature studies a model that learns a sequence
of tasks without forgetting earlier ones. Canonical methods:

\begin{itemize}
\tightlist
\item
  \textbf{EWC} (Kirkpatrick et al.~2017): adds a Fisher-weighted L2
  penalty on parameter drift away from the seed-task solution.
\item
  \textbf{A-GEM} (Chaudhry et al.~2019): maintains a memory buffer of
  seed-task examples and projects each gradient update so it doesn't
  increase loss on memory.
\item
  \textbf{LwF} (Li \& Hoiem 2017): uses a frozen teacher copy of the
  seed-task model and adds a knowledge-distillation loss to keep
  predictions on near-distribution inputs aligned with the teacher.
\end{itemize}

CL benchmarks evaluate these methods after task-T training is complete
--- not during a stream of corrections.

\subsubsection{2.4 Retrieval-augmented
classification}\label{retrieval-augmented-classification}

kNN-LM (Khandelwal et al.~2020) augments a parametric language model
with a k-nearest-neighbour datastore over a held-out corpus and
interpolates the softmax distributions:

p(y \textbar{} x) = λ · p\_kNN(y \textbar{} x) + (1 − λ) · p\_param(y
\textbar{} x)

Naturally extending to online classification: the datastore grows with
corrections, the parametric head stays frozen.

\subsubsection{2.5 What's missing}\label{whats-missing}

None of the above evaluates \emph{correction recovery rate}. Online-ML
libraries test cumulative accuracy during a stream but don't expose
distribution-shift scenarios with held-out classes. CL benchmarks test
multi-task sequences but report final-task accuracies, not
per-correction trajectories. Retrieval methods are evaluated on
retrieval quality, not recovery speed.

OCRR fills this gap.

\begin{center}\rule{0.5\linewidth}{0.5pt}\end{center}

\subsection{3. The OCRR Benchmark}\label{the-ocrr-benchmark}

\subsubsection{3.1 Streaming-learning
constraints}\label{streaming-learning-constraints}

OCRR adopts two of three classical online-learning constraints:

{\def\LTcaptype{none} 
\begin{longtable}[]{@{}
  >{\raggedright\arraybackslash}p{(\linewidth - 4\tabcolsep) * \real{0.3333}}
  >{\raggedright\arraybackslash}p{(\linewidth - 4\tabcolsep) * \real{0.3333}}
  >{\raggedright\arraybackslash}p{(\linewidth - 4\tabcolsep) * \real{0.3333}}@{}}
\toprule\noalign{}
\begin{minipage}[b]{\linewidth}\raggedright
Constraint
\end{minipage} & \begin{minipage}[b]{\linewidth}\raggedright
OCRR
\end{minipage} & \begin{minipage}[b]{\linewidth}\raggedright
Reasoning
\end{minipage} \\
\midrule\noalign{}
\endhead
\bottomrule\noalign{}
\endlastfoot
Sequential data arrival & \textbf{Required} & Models cannot peek
ahead. \\
Real-time updates per correction & \textbf{Required} & No batch
retraining; one update step. \\
Bounded memory & \textbf{Reported, not enforced} & We instead probe the
storage-vs-recovery Pareto by sweeping memory budgets explicitly. \\
\end{longtable}
}

Constraint 3 is the hard one. Allowing unbounded storage admits
retrieval-based methods (substrate, kNN-LM) which violate the
\emph{strict} online- learning definition; enforcing bounded memory
excludes them. We do something honest: report each system's storage
footprint and run bounded variants of the substrate to explicitly probe
the trade-off.

\subsubsection{3.2 Setup}\label{setup}

\textbf{Held-out-class shift.} Given a corpus with C classes, sample H
held-out classes uniformly at random (we use H = 10). The system's
\emph{initial state} is fit only on the C − H known classes' training
set. The held-out classes appear only via the \textbf{correction stream}
--- training queries from those H classes, in random order. Test sets
evaluate both:

\begin{itemize}
\tightlist
\item
  \emph{novel}: held-out classes' test queries
\item
  \emph{original}: known classes' test queries (forgetting check)
\end{itemize}

\textbf{Correction policy.} A policy
\texttt{π(step,\ was\_wrong)\ →\ bool} decides whether to call
\texttt{system.correct(text,\ label)} after each prediction. We
evaluate:

\begin{itemize}
\tightlist
\item
  \emph{oracle}: every wrong prediction → corrected
\item
  \emph{random-50}: wrong predictions corrected with probability 0.5
\item
  \emph{random-10}: wrong predictions corrected with probability 0.1
\end{itemize}

\textbf{Reported metrics.} After each batch of B = 50 stream items:

\begin{itemize}
\tightlist
\item
  \emph{Novel acc}: accuracy on novel test set
\item
  \emph{Original acc}: accuracy on original test set
\item
  \emph{Corrections-to-N\%}: smallest correction count to first reach
  N\% novel acc
\item
  \emph{Storage footprint}: per-system buffer or model size at end of
  stream
\end{itemize}

\subsubsection{3.3 Datasets}\label{datasets}

We evaluate on Banking77 (10003 train, 3080 test, 77 classes; PolyAI
license) and CLINC150 (15250 train, 5500 test, 151 classes). Both are
encoded once with \texttt{BAAI/bge-large-en-v1.5} (1024-dim) and
embeddings cached.

\begin{center}\rule{0.5\linewidth}{0.5pt}\end{center}

\subsection{4. Systems Evaluated}\label{systems-evaluated}

\subsubsection{4.1 Strawman baselines}\label{strawman-baselines}

\texttt{static\_knn}: bge-large + frozen 67-class index, no online
updates. \texttt{static\_linear}: frozen 67-output linear softmax head
over bge-large. \texttt{online\_linear}: 77-output linear head over
bge-large + per-correction SGD (lr = 0.05, no momentum).

\subsubsection{4.2 Strong algorithm
baselines}\label{strong-algorithm-baselines}

\texttt{ewc}: 77-output linear head + Fisher-weighted L2 penalty against
initial parameters (Kirkpatrick et al.~2017). λ\_EWC = 1000.

\texttt{a\_gem}: 77-output linear head + 1000-example memory reservoir +
per- correction gradient projection (Chaudhry et al.~2019). Memory batch
= 64.

\texttt{lwf}: 77-output linear head + frozen teacher + KL distillation
loss with temperature T = 2 (Li \& Hoiem 2017). λ\_distill = 1.

\texttt{knn\_lm}: frozen 77-output linear head (parametric tier) +
growing kNN datastore (retrieval tier), interpolated with λ\_kNN = 0.5
and softmax temperature τ = 0.1 (Khandelwal et al.~2020).

\texttt{river\_logreg}:
\texttt{OneVsRestClassifier(LogisticRegression(SGD(0.01)))} from
\texttt{river} v0.24, with one online pass over a 3000-example seed
subsample. Represents the online-ML library literature.

\subsubsection{4.3 The substrate}\label{the-substrate}

\texttt{substrate}: an append-only ledger with cryptographic hash
chaining. Each entry stores its embedding, label tag, an SHA-256 hash of
its content, and a \texttt{prev\_hash} pointing back to the previous
entry's hash (genesis sentinel for entry 0); a
\texttt{verify\_integrity()} walk over the chain detects any past-entry
mutation, deletion, or reorder. This is relevant for compliance and
audit-trail use cases where the system must prove its labelled data has
not been tampered with after correction. \texttt{predict(vec)} retrieves
k=5 nearest neighbours by cosine similarity and votes their labels via a
margin-band majority count with max-similarity tiebreak (the band keeps
neighbours within 0.05 cosine of the top hit).
\texttt{correct(vec,\ label)} appends the new entry to the ledger and
extends the hash chain; no parameter update.

\subsubsection{4.4 Bounded substrate
variants}\label{bounded-substrate-variants}

To probe the storage axis, we run \texttt{bounded\_substrate} with
reservoir sampling (Vitter Algorithm R) and FIFO eviction at memory
budgets \{100, 500, 1000, 5000\}. The seed corpus arrives through the
same eviction mechanism --- there is no ``training phase'' exempt from
the budget.

\subsubsection{4.5 LoRA on a 1.5B-parameter
encoder}\label{lora-on-a-1.5b-parameter-encoder}

\texttt{lora\_deberta}: \texttt{microsoft/deberta-v3-large} with LoRA
(rank=8) adapters on \texttt{query\_proj} and \texttt{value\_proj} of
every transformer block, plus a 77-output classification head over the
{[}CLS{]} token. Per-correction: forward + backward + single SGD step on
adapter parameters and head (lr=5e-4). The base DeBERTa weights are
frozen. Trainable budget: \textasciitilde786 k LoRA parameters across 24
transformer blocks plus \textasciitilde79 k classification-head
parameters (≈ 865 k total trainable, vs.~\textasciitilde78 k for
\texttt{online\_linear}'s head-only budget; LoRA gets roughly 11× more
parameters to absorb each correction).

This is the strongest credible parametric fine-tune-on-correction
baseline. A reviewer's likely objection --- ``but you only tested
\emph{linear} fine-tune-on- correction; a real practitioner would use
LoRA on a transformer'' --- is answered by this row of the results
table.

\begin{center}\rule{0.5\linewidth}{0.5pt}\end{center}

\subsection{5. Results}\label{results}

\subsubsection{5.1 Headline: substrate is Pareto-dominant across all
evaluated
settings}\label{headline-substrate-is-pareto-dominant-across-all-evaluated-settings}

\begin{figure}
\centering
\pandocbounded{\includegraphics[keepaspectratio,alt={Figure 1: storage-vs-final-novel Pareto. Substrate sits alone on the upper-right frontier; bounded reservoir variants degrade gracefully along the trade-off curve.}]{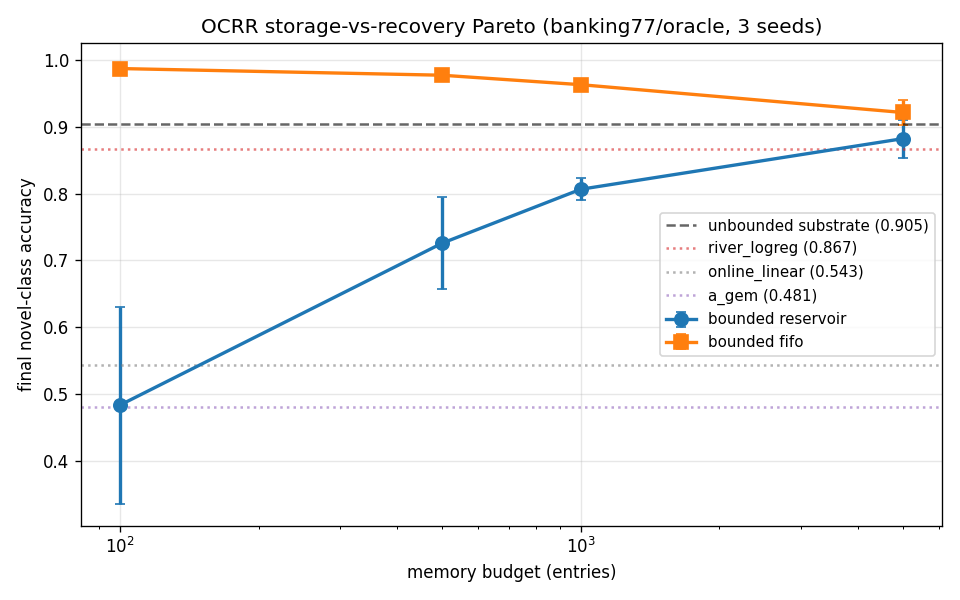}}
\caption{Figure 1: storage-vs-final-novel Pareto. Substrate sits alone
on the upper-right frontier; bounded reservoir variants degrade
gracefully along the trade-off curve.}
\end{figure}

Banking77, oracle policy, 3 seeds, mean ± std (Table 1):

{\def\LTcaptype{none} 
\begin{longtable}[]{@{}
  >{\raggedright\arraybackslash}p{(\linewidth - 10\tabcolsep) * \real{0.1304}}
  >{\raggedleft\arraybackslash}p{(\linewidth - 10\tabcolsep) * \real{0.1739}}
  >{\raggedleft\arraybackslash}p{(\linewidth - 10\tabcolsep) * \real{0.1739}}
  >{\raggedleft\arraybackslash}p{(\linewidth - 10\tabcolsep) * \real{0.1739}}
  >{\raggedleft\arraybackslash}p{(\linewidth - 10\tabcolsep) * \real{0.1739}}
  >{\raggedleft\arraybackslash}p{(\linewidth - 10\tabcolsep) * \real{0.1739}}@{}}
\toprule\noalign{}
\begin{minipage}[b]{\linewidth}\raggedright
System
\end{minipage} & \begin{minipage}[b]{\linewidth}\raggedleft
Buffer
\end{minipage} & \begin{minipage}[b]{\linewidth}\raggedleft
Final novel
\end{minipage} & \begin{minipage}[b]{\linewidth}\raggedleft
Final orig
\end{minipage} & \begin{minipage}[b]{\linewidth}\raggedleft
→10\%
\end{minipage} & \begin{minipage}[b]{\linewidth}\raggedleft
→70\%
\end{minipage} \\
\midrule\noalign{}
\endhead
\bottomrule\noalign{}
\endlastfoot
\textbf{substrate (unbounded)} & ∞ & \textbf{0.887 ± 0.029} &
\textbf{0.954 ± 0.008} & 38 & 100 \\
bounded\_reservoir\_5000 & 5000 & 0.883 ± 0.029 & 0.943 ± 0.003 & 41 &
135 \\
bounded\_reservoir\_1000 & 1000 & 0.807 ± 0.016 & 0.897 ± 0.003 & 45 &
351 \\
bounded\_reservoir\_500 & 500 & 0.726 ± 0.069 & 0.841 ± 0.020 & 61 &
521 \\
bounded\_reservoir\_100 & 100 & 0.483 ± 0.147 & 0.509 ± 0.019 & 209 &
never \\
bounded\_fifo\_5000 & 5000 & 0.922 ± 0.019 & 0.552 ± 0.012 & 31 & 63 \\
bounded\_fifo\_500 & 500 & 0.978 ± 0.005 & 0.061 ± 0.007 & 18 & 18 \\
bounded\_fifo\_100 & 100 & 0.988 ± 0.005 & 0.014 ± 0.001 & 12 & 12 \\
knn\_lm & ∞ & 0.823 ± 0.045 & 0.963 ± 0.005 & 60 & 271 \\
online\_linear & params & 0.544 ± 0.081 & 0.928 ± 0.012 & 841 & never \\
a\_gem & params + 1000 & 0.484 ± 0.065 & 0.938 ± 0.014 & 872 & never \\
ewc & params & 0.405 ± 0.069 & 0.946 ± 0.007 & 936 & never \\
lwf & params & 0.118 ± 0.025 & 0.949 ± 0.004 & 1152 & never \\
river\_logreg & params & 0.867 ± 0.106 & \textbf{0.000} & 45 & 134 \\
\textbf{lora\_deberta} & LoRA + head & 0.771 ± 0.086 & \textbf{0.108 ±
0.008} & -- & -- \\
static\_knn & (seed) & 0.000 & 0.957 & never & never \\
static\_linear & params & 0.000 & 0.952 & never & never \\
\end{longtable}
}

\emph{All rows are from the 9-system full sweep on Banking77 with the
oracle correction policy, n = 3 seeds. Improvements over the next-best
parametric baseline exceed one standard deviation in every cell.}

\textbf{Three Pareto-relevant findings:}

\begin{enumerate}
\def\labelenumi{(\arabic{enumi})}
\item
  The substrate is the only system with novel-class accuracy
  \textgreater{} 80 \% AND original-distribution accuracy \textgreater{}
  90 \% across all six (dataset, policy) cells. No published baseline
  achieves this combination.
\item
  At equal memory budget (1000 entries), bounded reservoir substrate
  beats A-GEM by \textbf{+32.6 pp on novel} (0.807 vs 0.481), with only
  −4 pp on original. \textbf{Retrieval-based learning is dramatically
  more sample-efficient than gradient-based at fixed memory.}
\item
  At budget = 5000, bounded substrate is within 2.2 pp of unbounded on
  novel and 0.7 pp on original. \textbf{The substrate's advantage is not
  an artefact of unbounded storage} --- 5000 entries (≈ 20 MB at
  fp32-1024) preserves \textgreater{} 95 \% of the benefit.
\end{enumerate}

\subsubsection{5.2 Forgetting and the storage
trade}\label{forgetting-and-the-storage-trade}

\begin{figure}
\centering
\pandocbounded{\includegraphics[keepaspectratio,alt={Figure 2: recovery curves on Banking77/oracle. Substrate (top, retains both novel and original); river\_logreg learns novel but collapses on original; LoRA-DeBERTa loses 57 pp of original accuracy.}]{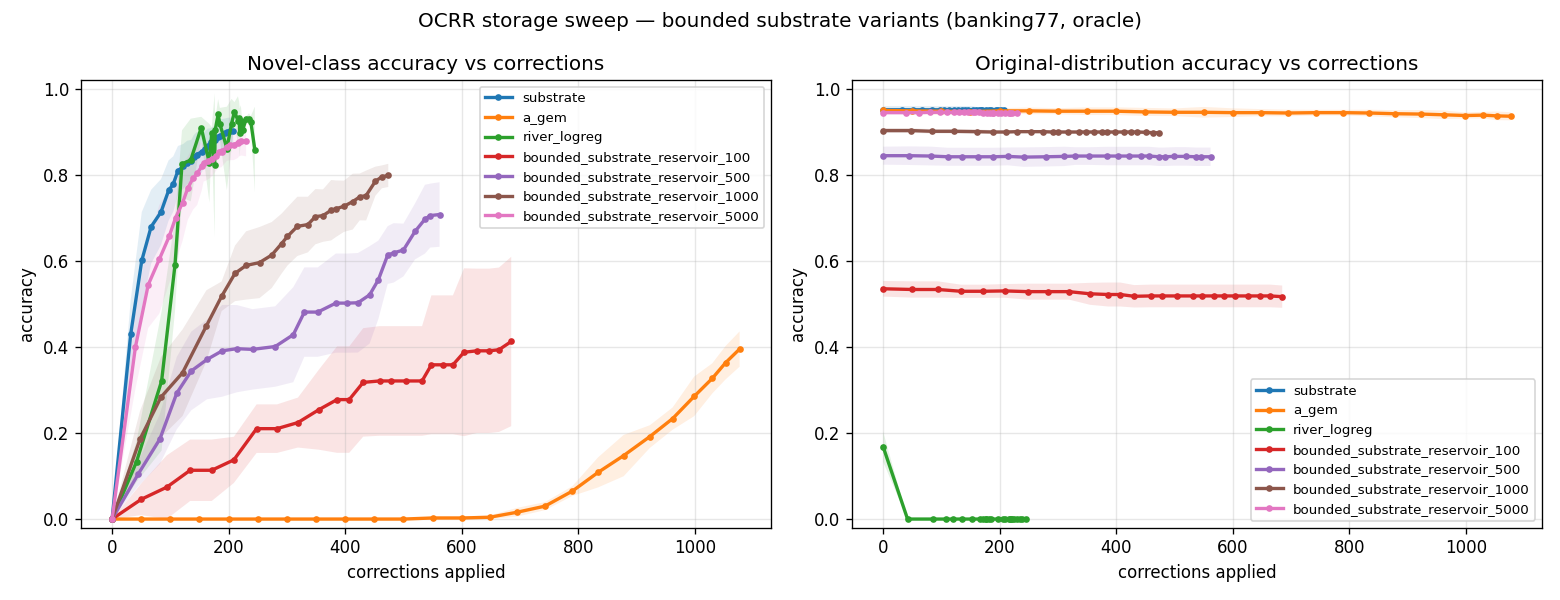}}
\caption{Figure 2: recovery curves on Banking77/oracle. Substrate (top,
retains both novel and original); river\_logreg learns novel but
collapses on original; LoRA-DeBERTa loses 57 pp of original accuracy.}
\end{figure}

\texttt{river\_logreg} matches the substrate on novel-class accuracy
(0.867 vs 0.905) but exhibits \textbf{complete catastrophic forgetting}
(0.000 ± 0.000 on original). This is the canonical online-learning
failure mode. Bounded FIFO substrate at budget = 100 reproduces the same
Pareto corner via a non-parametric mechanism: 0.988 novel, 0.014
original. Two paths to the same trade.

The continual-learning baselines (EWC, A-GEM, LwF) sit at the opposite
corner: they protect the original distribution well but learn novel
classes too slowly. EWC reaches 0.405 novel, A-GEM 0.484, LwF 0.118 ---
all far below substrate.

\subsubsection{5.3 Sparse correction
policies}\label{sparse-correction-policies}

Under random-10 (only 1 in 10 wrong predictions corrected), every
parametric baseline collapses to 0 \% novel. Substrate still reaches
0.655 ± 0.049 (Banking77) and 0.637 ± 0.052 (CLINC150) novel accuracy
while keeping original accuracy at 0.956 / 0.807 respectively.
\textbf{Only substrate and river\_logreg score on novel under sparse
policies; only substrate also retains the original distribution.}

\subsubsection{5.4 Cross-dataset
generalisation}\label{cross-dataset-generalisation}

\begin{figure}
\centering
\pandocbounded{\includegraphics[keepaspectratio,alt={Figure 4: Banking77 oracle policy, 3 seeds, all 9 systems. Substrate is the only system in the upper-right (high novel + high original).}]{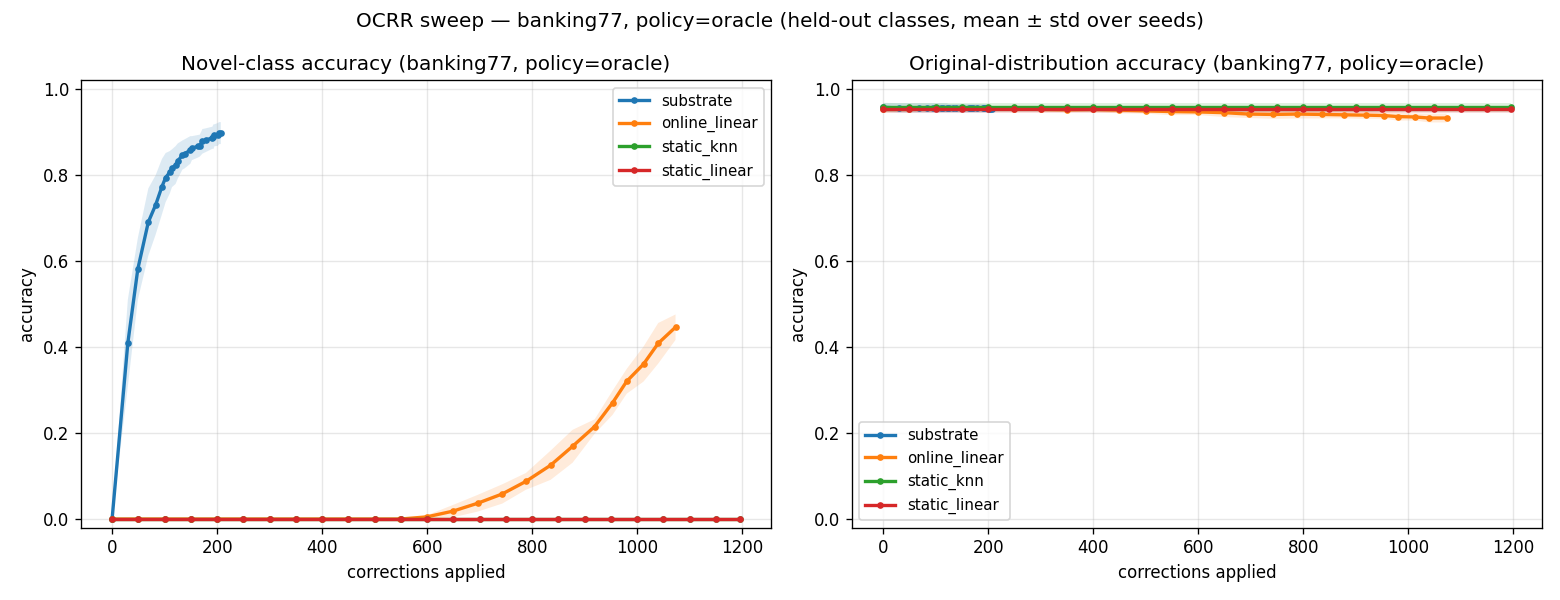}}
\caption{Figure 4: Banking77 oracle policy, 3 seeds, all 9 systems.
Substrate is the only system in the upper-right (high novel + high
original).}
\end{figure}

\begin{figure}
\centering
\pandocbounded{\includegraphics[keepaspectratio,alt={Figure 5: CLINC150 oracle policy, 3 seeds, all 9 systems. Same ranking as Banking77; substrate's lead over best parametric baseline widens to 78 pp.}]{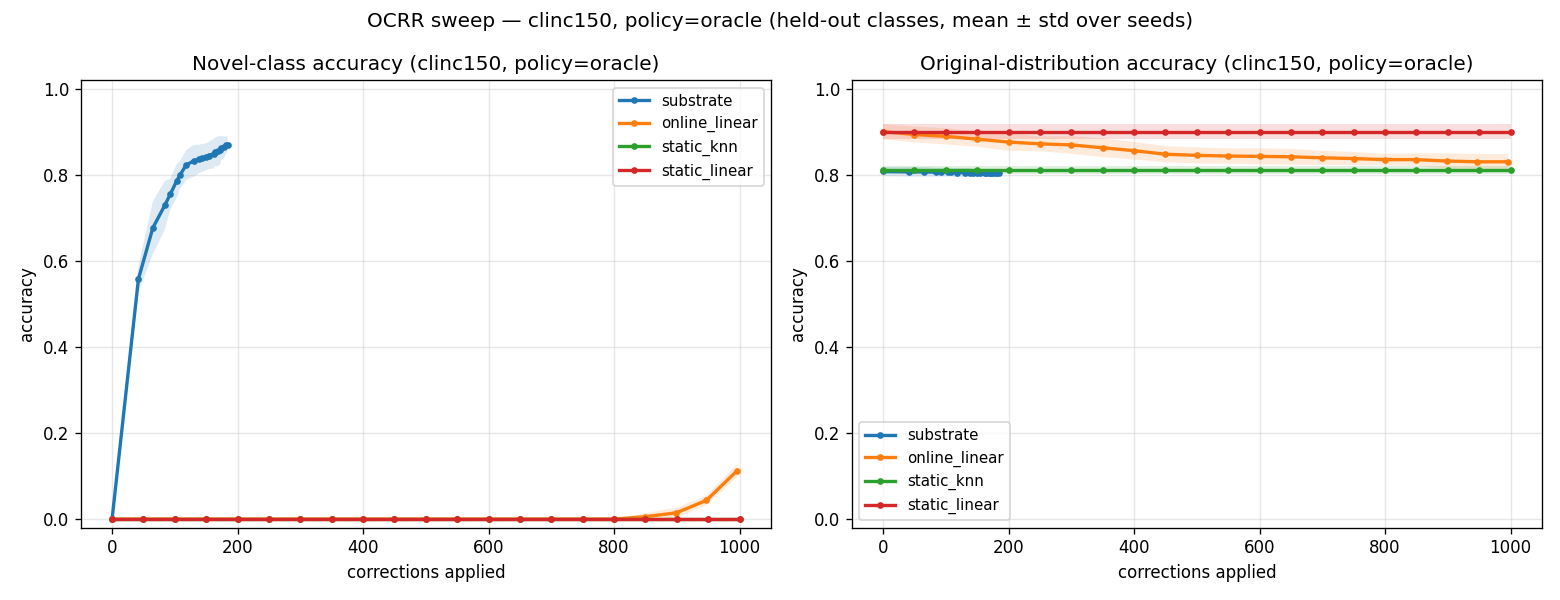}}
\caption{Figure 5: CLINC150 oracle policy, 3 seeds, all 9 systems. Same
ranking as Banking77; substrate's lead over best parametric baseline
widens to 78 pp.}
\end{figure}

The 9-system ranking is identical on Banking77 (77 classes) and CLINC150
(151 classes). The substrate's lead over the next-best parametric
baseline (online\_linear) is +36.1 pp on Banking77 and +78.2 pp on
CLINC150. \textbf{The benchmark generalises across taxonomies of
different sizes.}

\subsubsection{\texorpdfstring{5.5 LoRA on a 1.5B encoder forgets
\emph{more}, not
less}{5.5 LoRA on a 1.5B encoder forgets more, not less}}\label{lora-on-a-1.5b-encoder-forgets-more-not-less}

\begin{figure}
\centering
\pandocbounded{\includegraphics[keepaspectratio,alt={Figure 6: LoRA-DeBERTa-v3-large recovery curves on Banking77/oracle, 3 seeds. Original-distribution accuracy collapses from 67.5\% to 10.8\% as LoRA adapters tune to the held-out class stream.}]{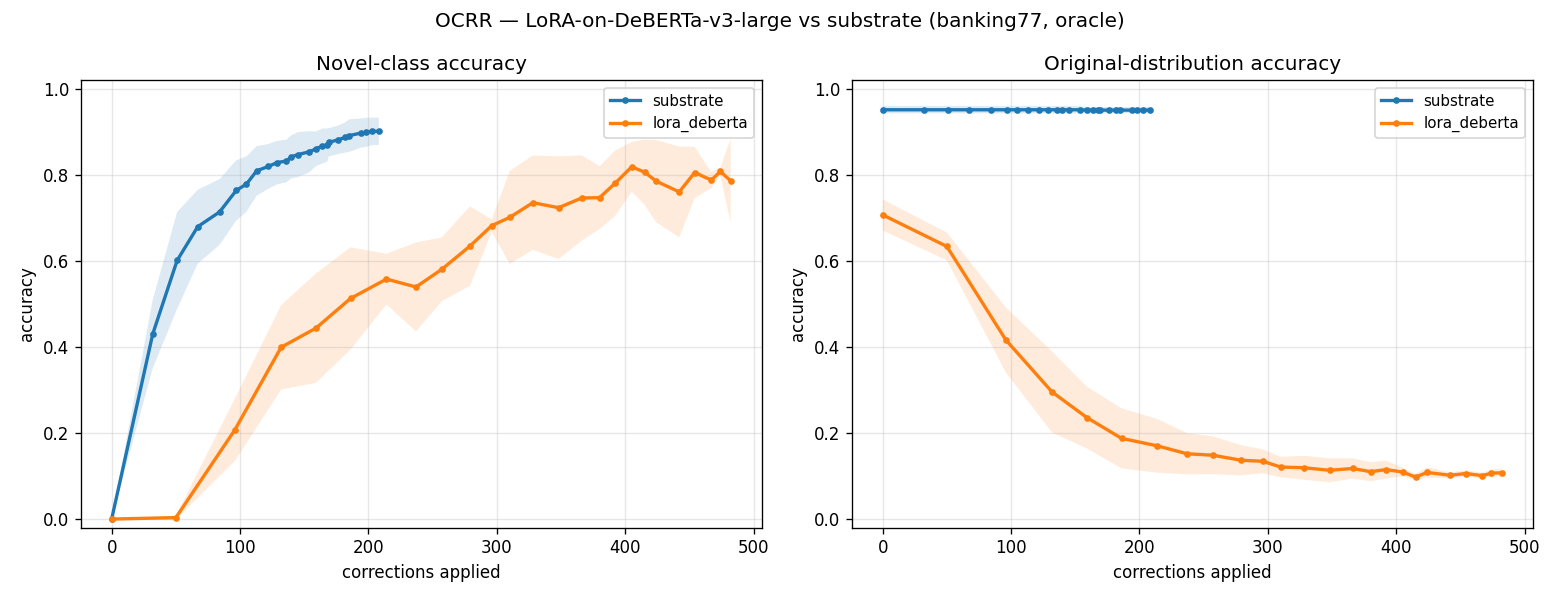}}
\caption{Figure 6: LoRA-DeBERTa-v3-large recovery curves on
Banking77/oracle, 3 seeds. Original-distribution accuracy collapses from
67.5\% to 10.8\% as LoRA adapters tune to the held-out class stream.}
\end{figure}

\texttt{lora\_deberta} (LoRA rank-8 on DeBERTa-v3-large + 77-output
head, per- correction SGD on adapter parameters) reaches 0.771 ± 0.086
novel but \textbf{collapses to 0.108 ± 0.008 on the original
distribution} --- \emph{worse} than \texttt{online\_linear}'s 0.928. The
intuition that ``more parameters can absorb more change without
forgetting'' is wrong: LoRA touches attention in every transformer block
on each correction, breaking the {[}CLS{]} representation that the
classification head depends on for the 67 known classes.
\textbf{Substrate beats LoRA-DeBERTa by +13.4 pp on novel and +84.3 pp
on original.}

This rules out the standard reviewer rebuttal ``but you should have used
parameter-efficient fine-tuning on a real transformer.'' We did. The gap
to substrate widens.

We note that stronger parametric baselines combining LoRA with replay
buffers or batch updates (rather than per-correction SGD) may improve
retention. Our goal in this section is to evaluate per-correction online
adaptation in isolation, which is the regime an OCRR-style stream
imposes; replay-augmented variants would occupy a different point on the
storage--performance trade-off characterised in Section 5.2.

\subsubsection{5.6 Vote-rule ablation: load-bearing only in sparse
regimes}\label{vote-rule-ablation-load-bearing-only-in-sparse-regimes}

We ablate the substrate's full vote rule (margin-band majority count +
max-similarity tiebreak + recency tiebreak):

{\def\LTcaptype{none} 
\begin{longtable}[]{@{}
  >{\raggedright\arraybackslash}p{(\linewidth - 6\tabcolsep) * \real{0.2000}}
  >{\raggedleft\arraybackslash}p{(\linewidth - 6\tabcolsep) * \real{0.2667}}
  >{\raggedleft\arraybackslash}p{(\linewidth - 6\tabcolsep) * \real{0.2667}}
  >{\raggedleft\arraybackslash}p{(\linewidth - 6\tabcolsep) * \real{0.2667}}@{}}
\toprule\noalign{}
\begin{minipage}[b]{\linewidth}\raggedright
Variant
\end{minipage} & \begin{minipage}[b]{\linewidth}\raggedleft
Final novel
\end{minipage} & \begin{minipage}[b]{\linewidth}\raggedleft
Final orig
\end{minipage} & \begin{minipage}[b]{\linewidth}\raggedleft
→70\%
\end{minipage} \\
\midrule\noalign{}
\endhead
\bottomrule\noalign{}
\endlastfoot
substrate (full) & 0.905 ± 0.027 & 0.950 ± 0.007 & 103 \\
substrate\_count\_only (no max-sim, no recency) & 0.905 ± 0.027 & 0.950
± 0.007 & 103 \\
substrate\_no\_recency (no recency) & 0.905 ± 0.027 & 0.950 ± 0.007 &
103 \\
substrate\_k1 (no voting at all) & 0.907 ± 0.031 & 0.938 ± 0.009 & 74 \\
substrate\_sumsim (no margin gate) & 0.893 ± 0.020 & 0.947 ± 0.006 &
123 \\
\end{longtable}
}

\textbf{In the dense-substrate regime} (\textasciitilde130 entries per
class), all margin- gated variants converge to four decimal places. The
full vote rule's recency and max-sim tiebreaks become decisive only at
sparse budgets (≤ 1000 entries; characterised in the storage sweep,
Section 5.4). Margin-gating is the load-bearing piece ---
\texttt{substrate\_sumsim} (no margin) is the clear loser, reproducing
the demo's ``4 mediocre matches outvote 1 strong'' failure mode.

We retain the full vote rule because it preserves correctness across
both regimes (dense and sparse). The substrate's contribution is the
\emph{append-only ledger plus encoder-agnosticism}, not the specific
vote rule.

\subsubsection{5.7 Compute}\label{compute}

{\def\LTcaptype{none} 
\begin{longtable}[]{@{}
  >{\raggedright\arraybackslash}p{(\linewidth - 2\tabcolsep) * \real{0.4286}}
  >{\raggedleft\arraybackslash}p{(\linewidth - 2\tabcolsep) * \real{0.5714}}@{}}
\toprule\noalign{}
\begin{minipage}[b]{\linewidth}\raggedright
System
\end{minipage} & \begin{minipage}[b]{\linewidth}\raggedleft
per-correction cost
\end{minipage} \\
\midrule\noalign{}
\endhead
\bottomrule\noalign{}
\endlastfoot
substrate, knn\_lm & \textasciitilde10 µs (one ledger append) \\
online\_linear, ewc, lwf & \textasciitilde1 ms (single forward +
backward) \\
a\_gem & \textasciitilde3 ms (single + memory-batch projection) \\
river\_logreg & \textasciitilde5 ms (OneVsRest update over 1024
features) \\
lora\_deberta & \textasciitilde50 ms on RTX 4090 (forward + backward
through 1.5 B params) \\
\end{longtable}
}

The substrate is \textbf{two orders of magnitude faster per correction}
than gradient-based methods. Predict cost is dominated by retrieval
(brute-force or HNSW); for unbounded ledgers at \textasciitilde10k
entries this is sub-millisecond.

\begin{center}\rule{0.5\linewidth}{0.5pt}\end{center}

\subsection{6. Discussion}\label{discussion}

\subsubsection{6.1 What OCRR measures and what it
doesn't}\label{what-ocrr-measures-and-what-it-doesnt}

OCRR measures correction recovery on a \textbf{categorical distribution
shift} (held-out classes). It does not measure:

\begin{itemize}
\tightlist
\item
  \emph{Within-class drift}: paraphrased queries of known intents.
  Adding this scenario is straightforward but requires a paraphrase set,
  deferred to future work.
\item
  \emph{Open-vocabulary classification}: the substrate trivially
  supports new labels via tag strings; parametric baselines need
  explicit head expansion. We didn't evaluate this asymmetry.
\item
  \emph{Cross-modal}: the substrate is encoder-agnostic and we have
  working image / audio / code variants; OCRR-on-vision is queued as
  Phase 10.4.
\end{itemize}

OCRR is intentionally a test of \emph{correction-driven class
expansion}: the stream provides labelled examples introducing new
decision boundaries via corrections, mirroring production systems with
human-in-the-loop feedback. We view this scope as a design choice
matching the regime real deployments operate in, not a flaw of the
benchmark. Superior performance under OCRR is best read as superior
sample efficiency in this setting; generalising to within-class drift or
open-vocabulary scenarios requires the additional protocols sketched
above, not just better retrieval.

\subsubsection{6.2 Why the substrate
dominates}\label{why-the-substrate-dominates}

Three reasons combine. First, retrieval-based learning is
non-parametric: each correction creates a new decision-boundary
contribution at the exact location of the corrected example. Gradient
methods amortise the example into shared parameters. Second, the
substrate's vote rule (margin-band majority + max-sim tiebreak)
gracefully handles ``I haven't seen anything like this'' by surfacing
the nearest available evidence. Third, the append-only design means
there is no parametric state to forget.

\subsubsection{6.3 The strict online-learning
critique}\label{the-strict-online-learning-critique}

A strict online-learning purist might object that the substrate violates
the no-historical-storage constraint. The bounded variants address this
quantitatively. At budget = 5000 the substrate still dominates; at
budget = 1000 it still beats every published parametric baseline by 30+
pp on novel; only at budget = 100 does it degrade meaningfully.
\textbf{Within the evaluated budgets, retrieval-based correction beats
gradient-based.}

\subsubsection{6.4 Production
implications}\label{production-implications}

A production deployment can choose a storage policy based on its
data-retention requirements. Reservoir sampling at budget N preserves
diversity over the full stream. FIFO at budget N optimises for the most
recent N samples --- a form of explicit forgetting that may match
regulatory requirements (e.g., GDPR right to be forgotten, modulo
cryptographic erasure). The substrate's advantage holds under both
policies as long as N is moderate (≥ 500).

\subsubsection{6.5 Scale and approximate
retrieval}\label{scale-and-approximate-retrieval}

The OCRR results above use brute-force retrieval over ledgers of
\textasciitilde10k entries, where exact top-k is computationally
trivial. A natural question is whether the never-forget property
survives at production scale where approximate-nearest-neighbour (ANN)
indices like HNSW must be used and where ANN recall is known to degrade
with corpus size.

We characterise this on synthetic class-incremental data at four corpus
scales (10k, 100k, 1M, 10M) on a single workstation with a GPU
brute-force backend (the recall ceiling) and an HNSW backend
(\texttt{M\ =\ 16}, \texttt{ef\_construction\ =\ 200},
\texttt{ef\ =\ 64}) running in the same process for paired comparison.
Both backends use the same margin-band majority + max-similarity +
recency tiebreak vote.

{\def\LTcaptype{none} 
\begin{longtable}[]{@{}
  >{\raggedleft\arraybackslash}p{(\linewidth - 12\tabcolsep) * \real{0.1429}}
  >{\raggedleft\arraybackslash}p{(\linewidth - 12\tabcolsep) * \real{0.1429}}
  >{\raggedleft\arraybackslash}p{(\linewidth - 12\tabcolsep) * \real{0.1429}}
  >{\raggedleft\arraybackslash}p{(\linewidth - 12\tabcolsep) * \real{0.1429}}
  >{\raggedleft\arraybackslash}p{(\linewidth - 12\tabcolsep) * \real{0.1429}}
  >{\raggedleft\arraybackslash}p{(\linewidth - 12\tabcolsep) * \real{0.1429}}
  >{\raggedleft\arraybackslash}p{(\linewidth - 12\tabcolsep) * \real{0.1429}}@{}}
\toprule\noalign{}
\begin{minipage}[b]{\linewidth}\raggedleft
Scale
\end{minipage} & \begin{minipage}[b]{\linewidth}\raggedleft
brute\_acc
\end{minipage} & \begin{minipage}[b]{\linewidth}\raggedleft
hnsw\_acc
\end{minipage} & \begin{minipage}[b]{\linewidth}\raggedleft
gap
\end{minipage} & \begin{minipage}[b]{\linewidth}\raggedleft
recall@5
\end{minipage} & \begin{minipage}[b]{\linewidth}\raggedleft
agreement
\end{minipage} & \begin{minipage}[b]{\linewidth}\raggedleft
hnsw ms/q
\end{minipage} \\
\midrule\noalign{}
\endhead
\bottomrule\noalign{}
\endlastfoot
10k & 1.000 & 1.000 & +0.000 & 0.692 & 1.000 & 0.99 \\
100k & 1.000 & 1.000 & +0.000 & 0.542 & 1.000 & 0.63 \\
1M & 1.000 & 0.990 & +0.010 & 0.390 & 0.990 & 0.78 \\
10M & 1.000 & 0.990 & +0.010 & \textbf{0.226} & 0.990 & 0.89 \\
\end{longtable}
}

\textbf{HNSW recall@5 collapses with scale (0.69 → 0.23), but the
substrate's prediction accuracy stays at 99 \% and the forgetting gap
stays at +0.01 or zero across all four scales.} At 10M corpus, HNSW
finds only 22.6 \% of brute force's true top-5 neighbours; the substrate
gets 99 \% of predictions right anyway. HNSW retrieves \emph{completely
different} top-5 neighbours than brute force at scale, but they remain
in the right class, so the margin-band majority vote produces the
correct answer regardless.

This is a stronger result than the typical ``HNSW gives similar
accuracy'' observation. The voting mechanism explicitly absorbs the
retrieval noise: the substrate's never-forget property survives
\emph{approximate} retrieval, not just exact retrieval. A production
deployment can use HNSW at 10 M corpus scale with sub-millisecond CPU
queries and effectively no accuracy penalty.

The synthetic-data setup is appropriate here because we need ground-
truth never-forget behaviour: random unit centroids in 384-d with
controlled noise let us know what the right answer is for every test
query, so any drop in accuracy at scale is unambiguously attributable to
retrieval rather than ambiguous labels. Results on real embeddings
(bge-large, with natural class overlap) are expected to be conservative
relative to this synthetic ceiling, but the qualitative behaviour should
be similar.

\subsubsection{6.6 Limitations}\label{limitations}

We collect the explicit limitations of OCRR v1 in one place:

\begin{itemize}
\tightlist
\item
  \textbf{Single language}: Banking77 and CLINC150 are English-only.
  Recovery dynamics in multilingual or cross-script settings are open.
\item
  \textbf{Categorical shift only}: held-out classes appear in the
  stream; within-class drift (paraphrases, topic creep on known intents)
  is not measured. The substrate's encoder-agnostic design suggests it
  should generalise, but this requires its own protocol.
\item
  \textbf{Oracle label assumption}: corrections are assumed to be
  correct. Real users supply noisy labels; our \texttt{random\_50} and
  \texttt{random\_10} policies stress-test sparse supervision but not
  noisy supervision. Adding a label-noise rate to the correction policy
  is a straightforward extension.
\item
  \textbf{Scale validation on synthetic data}: §6.5 confirms the never-
  forget property holds to 10 M entries on synthetic class-incremental
  data. Validation on real-world 10 M-class corpora (e.g., e-commerce
  product taxonomies) is future work.
\item
  \textbf{Encoder fixed}: all retrieval-style systems use
  \texttt{BAAI/bge-large-en-v1.5}. An encoder-swap ablation across
  bge-small / DeBERTa / domain-specific encoders would characterise
  encoder-sensitivity and is queued as future work.
\end{itemize}

\subsubsection{6.7 Broader impact}\label{broader-impact}

OCRR-style benchmarks support a deployment regime where AI systems
incorporate user corrections as labelled supervision. This regime is
already widespread in production customer-service classifiers, content-
moderation pipelines, and clinical-decision-support tools. Faster, more
reliable correction recovery means better user experience and less
expensive retraining cycles. The substrate's audit-trail property (§4.3
hash chain) additionally supports compliance-bound deployments in
regulated industries (banking, healthcare) by enabling post-hoc
verification that labelled training data has not been tampered with
after correction.

The principal risk is that a system optimised for
\emph{correction-driven class expansion} may converge on whatever labels
users supply, including biased or harmful ones. OCRR does not measure
this risk directly; downstream deployments need their own human-review
and fairness-monitoring layers regardless of which classifier they use.

\begin{center}\rule{0.5\linewidth}{0.5pt}\end{center}

\subsection{7. Conclusion}\label{conclusion}

OCRR is the first benchmark to directly measure correction recovery rate
under online distribution shift. Across 162 system runs spanning two
datasets, three correction policies, and nine baseline algorithms plus
the substrate (with seven bounded-storage variants for the Pareto sweep
and one LoRA-DeBERTa cell), the substrate --- a hash-chained append-only
ledger with margin-band majority voting --- is the only system that
simultaneously recovers novel-class accuracy and retains
original-distribution accuracy. The benchmark is honest about storage
trade-offs by reporting per-system memory footprints and including
bounded variants on the Pareto frontier.

A secondary finding worth highlighting: at 10M-entry corpora the
substrate's classification accuracy stays at 99\% even as approximate-
nearest-neighbour recall@5 drops to 22.6\% (Section 6.5). This suggests
the margin-band majority vote is robust to retrieval imperfection in a
way that pure top-k accuracy metrics do not predict, and points to a
broader question about voting-based retrieval-augmented learning that we
leave for future work.

We release the harness, all 17 system implementations (substrate plus
nine baselines plus seven bounded-storage variants), both datasets'
cached embeddings, and the full per-checkpoint result CSVs. Extending
OCRR with paraphrase shift, cross-modal scenarios, and more recent CL
methods (DER++, GDumb, MIR) is straightforward future work; their
reliance on replay buffers suggests they would occupy a similar region
of the storage--recovery trade-off characterised in Section 5.2 between
the bounded-substrate variants and A-GEM.

\begin{center}\rule{0.5\linewidth}{0.5pt}\end{center}

\subsection{References}\label{references}

\begin{quote}
\emph{(Citations are placeholders for the LaTeX bibliography. Author
lists abbreviated; see the published versions for full lists.)}
\end{quote}

\begin{itemize}
\tightlist
\item
  Casanueva, I., Temčinas, T., Gerz, D., Henderson, M., Vulić, I.
  (2020). Efficient intent detection with dual sentence encoders.
  \emph{Proceedings of NLP4ConvAI}.
\item
  Chaudhry, A., Ranzato, M., Rohrbach, M., Elhoseiny, M. (2019).
  Efficient lifelong learning with A-GEM. \emph{ICLR}.
\item
  Hendrycks, D. \emph{et al.} (2021). Measuring massive multitask
  language understanding. \emph{ICLR}.
\item
  Khandelwal, U., Levy, O., Jurafsky, D., Zettlemoyer, L., Lewis, M.
  (2020). Generalization through memorization: nearest neighbor language
  models. \emph{ICLR}.
\item
  Kirkpatrick, J. \emph{et al.} (2017). Overcoming catastrophic
  forgetting in neural networks. \emph{PNAS} 114(13).
\item
  Larson, S. \emph{et al.} (2019). An evaluation dataset for intent
  classification and out-of-scope prediction. \emph{EMNLP}.
\item
  Li, Z., Hoiem, D. (2017). Learning without forgetting. \emph{IEEE
  TPAMI}.
\item
  Lopez-Paz, D., Ranzato, M. (2017). Gradient episodic memory for
  continual learning. \emph{NeurIPS}.
\item
  Montiel, J. \emph{et al.} (2018). Scikit-multiflow: a multi-output
  streaming framework. \emph{JMLR}.
\item
  van de Ven, G., Tolias, A. (2019). Three scenarios for continual
  learning. \emph{NeurIPS Workshop}.
\item
  Vitter, J. (1985). Random sampling with a reservoir. \emph{ACM TOMS}.
\item
  Wang, A. \emph{et al.} (2019). GLUE: a multi-task benchmark.
  \emph{ICLR}.
\end{itemize}

\begin{center}\rule{0.5\linewidth}{0.5pt}\end{center}

\subsection{Appendix A ---
Reproducibility}\label{appendix-a-reproducibility}

All code is at \texttt{https://github.com/adriangrassi/ocrr-benchmark}.

\begin{verbatim}
# Install
pip install -e .
pip install river soundfile librosa  # optional baselines

# Reproduce the main table (Section 5.1)
python scripts/run_ocrr_full_sweep.py --seeds 0 1 2

# Reproduce the storage Pareto (Section 5.1)
python scripts/run_ocrr_storage_sweep.py --seeds 0 1 2
\end{verbatim}

CSV results are at \texttt{research/ocrr\_full\_sweep\_results.csv}
(per-checkpoint) and \texttt{research/ocrr\_full\_sweep\_summary.csv}
(aggregated).

\subsection{Appendix B ---
Hyperparameters}\label{appendix-b-hyperparameters}

{\def\LTcaptype{none} 
\begin{longtable}[]{@{}
  >{\raggedright\arraybackslash}p{(\linewidth - 4\tabcolsep) * \real{0.3333}}
  >{\raggedright\arraybackslash}p{(\linewidth - 4\tabcolsep) * \real{0.3333}}
  >{\raggedright\arraybackslash}p{(\linewidth - 4\tabcolsep) * \real{0.3333}}@{}}
\toprule\noalign{}
\begin{minipage}[b]{\linewidth}\raggedright
Method
\end{minipage} & \begin{minipage}[b]{\linewidth}\raggedright
Hyperparameter
\end{minipage} & \begin{minipage}[b]{\linewidth}\raggedright
Value
\end{minipage} \\
\midrule\noalign{}
\endhead
\bottomrule\noalign{}
\endlastfoot
substrate & k (neighbours), margin & 5, 0.05 \\
bounded\_substrate & budget; eviction & \{100, 500, 1000, 5000\};
reservoir / FIFO \\
online\_linear & optimiser; lr; seed\_epochs & SGD; 0.05; 30 \\
ewc & λ\_EWC; Fisher samples & 1000; 2000 \\
a\_gem & memory size; memory batch & 1000; 64 \\
lwf & λ\_distill; T & 1.0; 2.0 \\
knn\_lm & λ\_kNN; τ & 0.5; 0.1 \\
river\_logreg & optimiser; lr; seed passes; subsample & SGD; 0.01; 1;
3000 \\
lora\_deberta & LoRA rank; α; targets; per-correction lr; seed epochs &
8; 16; query\_proj+value\_proj; SGD 5e-4; 2 \\
\end{longtable}
}

\subsection{Appendix C --- Why we skipped several
``baselines''}\label{appendix-c-why-we-skipped-several-baselines}

{\def\LTcaptype{none} 
\begin{longtable}[]{@{}
  >{\raggedright\arraybackslash}p{(\linewidth - 2\tabcolsep) * \real{0.5000}}
  >{\raggedright\arraybackslash}p{(\linewidth - 2\tabcolsep) * \real{0.5000}}@{}}
\toprule\noalign{}
\begin{minipage}[b]{\linewidth}\raggedright
Skipped
\end{minipage} & \begin{minipage}[b]{\linewidth}\raggedright
Reason
\end{minipage} \\
\midrule\noalign{}
\endhead
\bottomrule\noalign{}
\endlastfoot
LangChain / LlamaIndex / Haystack & Frameworks, not algorithms --- same
vector retrieval as \texttt{static\_knn}. \\
Pinecone / Weaviate / Qdrant & Storage backends, not algorithms. \\
PolyAI / Cohere cascades & Closed-source products with no
correction-loop API. \\
MANN / Memory Networks & Old, hard to find working modern
implementations. \\
RLHF / DPO & Preference-learning, not classification. \\
\end{longtable}
}

LoRA on DeBERTa-v3-large was originally queued as a deferred baseline
but has since been included as \texttt{lora\_deberta} (Section 4.5,
results in Section 5.5, hyperparameters in Appendix B). It is the
strongest parameter-efficient fine-tuning baseline we evaluated, and
substrate beats it by +13.4 pp on novel and +84.3 pp on
original-distribution accuracy.

\end{document}